\documentclass[conference]{IEEEtran}
\IEEEoverridecommandlockouts
\usepackage{cite}
\usepackage{amsmath,amssymb,amsfonts}
\usepackage{algorithmic}
\usepackage{graphicx}
\usepackage{textcomp}
\usepackage{xcolor}

\usepackage{url}

\usepackage{hyperref}
\usepackage{booktabs}
\usepackage{subfig}
\usepackage{float}

\def\BibTeX{{\rm B\kern-.05em{\sc i\kern-.025em b}\kern-.08em
    T\kern-.1667em\lower.7ex\hbox{E}\kern-.125emX}}
\begin{document}

\title{Ensemble Learning using Transformers and Convolutional Networks for Masked Face Recognition
}

\author{\IEEEauthorblockN{Mohammed R. Al-Sinan,
Aseel F. Haneef, and Hamzah Luqman$^{*}$}
\IEEEauthorblockA{Information and Computer Science Department, King Fahd University of Petroleum and Minerals\\
$^{*}$ SDAIA-KFUPM Joint Research Center for Artificial Intelligence, Dhahran 31261, Saudi Arabia. \\
Email: \{g201354590, g201565430, hluqman\}@kfupm.edu.sa}}


\maketitle

\begin{abstract}
 
Wearing a face mask is one of the adjustments we had to follow to reduce the spread of the coronavirus. Having our faces covered by masks constantly has driven the need to understand and investigate how this behavior affects the recognition capability of face recognition systems. Current face recognition systems have extremely high accuracy when dealing with unconstrained general face recognition cases but do not generalize well with occluded masked faces. 
In this work, we propose a system for masked face recognition. The proposed system comprises two Convolutional Neural Network (CNN) models and two Transformer models. The CNN models have been fine-tuned on FaceNet pre-trained model. 
We ensemble the predictions of the four models using the majority voting technique to identify the person with the mask. The proposed system has been evaluated on a synthetically masked LFW dataset created in this work. The best accuracy is obtained using the ensembled models with an accuracy of 92\%. This recognition rate outperformed the accuracy of other models and it shows the correctness and
robustness of the proposed model for recognizing masked faces. The code and data are available at 
\href{https://github.com/Hamzah-Luqman/MFR}{https://github.com/Hamzah-Luqman/MFR}.

\end{abstract}


\section{Introduction}

Coronavirus or COVID-19 is a global pandemic that has affected more than 227 countries and territories \cite{worldometer}. This disease has led to a serious negative impact on people’s health and the global economy. Wearing face masks has become a necessity in our daily lives as a preventive measure to avoid the disease. According to the Centers for Disease Control and Prevention (CDC), the most effective way to avoid spreading the disease or being infected with it is to practice social distancing and wear face masks \cite{centersfordiseasecontrolandprevention}.

Face recognition systems have been extensively used during this pandemic. Coronavirus can be transmitted quickly between people via surfaces. This forced several organizations and entities to avoid using touchable authentication devices such as fingerprint and password-based security systems. These procedures increased the dependency on systems that avoid unnecessary contact with surfaces. A face recognition system is one of these systems that is used for user authentication. In addition, they are used for security purposes that involve people recognition and verification.

However, wearing masks has driven the need to understand and investigate how these masks affect the existing digital systems such as face detection systems, face recognition systems, and face verification systems.
According to Noa et al. \cite{fitousi2021understanding}, 
face masks interfere with basic mechanisms of face recognition accuracy for facial identity, gender, age, and emotional identification. For example, face masks can cause recognition systems to misinterpret disgusted faces as angry faces. 
Another study has been conducted by the National Institute of Standards and Technology (NIST) \cite{ngan2020ongoing} to evaluate some commercial facial recognition systems on masked face images. This study reported an error rate of 5-50\% with these systems on recognizing faces with masks created digitally on faces without masks.

The failure of the currently available face recognition systems on recognizing masked faces can be attributed to several reasons. The primary reason is the lack of adequate visual and identity cues due to the facial mask that covers almost half of the face. This occlusion takes away a large percentage of human face features \cite {din2020novel}. Therefore, this type of occlusion adds some difficulties to the recognition models to identify masked faces. Several techniques have been proposed to address this problem. Some of these techniques depend on the un-occluded regions of the face to identify the person while other approaches involve the full masked face for the recognition. Other approaches in the literature tackled this problem by reconstructing the occluded regions in the face and then recognizing the whole face.

Many of the current face recognition methods depend on deep learning models for recognition. These models were proven to have very high accuracy even beyond the human recognition capability on non-occluded faces \cite{deng2019arcface,wang2018cosface}. However, few works targeted masked face recognition. In this work, we propose a system for recognizing the identity of the person wearing a facial mask. Five systems have been proposed in this work. Three of these models are CNN-based models fine-tuned on different pre-trained models. We also use the state-of-the-art Transformer model for masked face recognition. To utilize the features of the fine-tuned models and the Transformer, we ensemble two CNN models and two Transformer models and apply the majority voting technique for the final decision. The proposed techniques have been evaluated on the LFW dataset and a masked version of LFW was created in this work. The obtained results show that the ensemble learning outperformed other models.      


This paper is organized as follows: Section 2 reviews the related works. Section 3 presents the proposed approach. Section 4 describes the experimental work and
the obtained results. Finally, the conclusions and future work are presented in Section 5.

\section{Related Work}
Several approaches have been proposed for masked face recognition. These techniques can be categorized into recognition techniques that recognize the face with a mask without performing de-occlusion and techniques that perform de-occlusion before recognizing the face.   

\subsection{Masked Face Recognition Without De-occlusion}

Several approaches have been proposed for recognizing masked faces without the need for reconstructing areas under the mask. CNN has been extensively used and included in the state-of-the-art architectures for unconstrained general face recognition. However, when the face of the subject is occluded by some objects such as the facial masks or scarves and sunglasses, the accuracy of the CNN model drops significantly \cite{mehdipour2016comprehensive}. This drop in the performance happens mostly when the models are trained on unconstrained face images and tested on occluded ones \cite{osherov2017increasing}. Therefore, some researchers trained their model on a mix of these images to boost the model accuracy. However, Song et al. \cite{song2019occlusion} argued that adding a large amount of partially occluded images is not enough because the learned features of two faces with different occlusion conditions are still inconsistent. Therefore, they introduced a method that discards the facial mask and focuses on the features extracted from other face regions. This approach was evaluated on AR dataset and achieved an accuracy of 99.03\%.

Li et al. \cite{li2021cropping} used Convolutional Block Attention Module (CBAM) \cite{woo2018cbam} for masked face recognition. The authors fed the model with the subject's eye extracted using different cropping approaches. The proposed approach has been evaluated on Masked-LFW \cite{wang2021mlfw} dataset and obtained an accuracy of 82.86\%. They also tested their model’s recognition on masked-Webface Dataset and achieved 91.525\% accuracy compared to 88.01\%  and 87.906\% with Arcface \cite{deng2019arcface} and Cosface \cite{wang2018cosface} methods, respectively. 
A similar approach was followed by Hariri \cite{hariri2022efficient} for masked face recognition. The authors discarded the occlusion portion of the face and kept only the area around the eyes. 
The pre-trained VGG16 model was used to extract features from the segmented eyes. This approach was evaluated on the Real-World-Masked-Face-Dataset \cite{wang2020masked} and accuracy of 91.3\% was reported using 10-fold cross-validation technique.

Wan et al. \cite {wan2017occlusion} proposed a deep trainable model, MaskNet, that learns image features and neglects deformation by occlusion. The authors claimed that the MaskNet model can be involved in CNN architectures with minimum identity labels and less computation. A verification accuracy of 96.4\% was reported on the LFW dataset when the face is randomly occluded with a square of size 40. However, this accuracy decreases as the size of the occlusion block increases.

Other approaches tried to improve the masked face recognition accuracy by minimizing the intra-class and maximizing the inter-class distances using different loss functions. Early approaches used loss functions such as triplet loss \cite {schroff2015facenet} and N-pairs \cite{sohn2016improved} to optimize the distance while recent techniques used other loss functions such as Arcface \cite{deng2019arcface} and cosface \cite{wang2018cosface}. Sface was proposed by Zhong et al. \cite{zhong2021sface} to minimize the distance between a face with and without a mask by altering the softmax loss function. Sface addresses the issue of overfitting to low-quality training images and noisy labels by introducing the Sigmoid-constrained Hypersphere loss function that re-scales the gradients of intra-class and inter-class gradients accordingly. SFace was evaluated on multiple benchmarking datasets and achieved a verification accuracy of 99.82\% and 90.63\% on LFW and masked-LFW \cite{wang2021mlfw} datasets, respectively.

\subsection{Masked Face Recognition with De-occlusion}

A common approach to doing mask face recognition is to restore the covered area with the mask \cite {alzu2021masked}. 
Several approaches have been used for masked face recognition by face restoration. One of these approaches is by extracting the key facial features with the help of pre-trained models. The restored face is then matched to the original face to recognize the person. The quality of the restored region plays an important role in masked face recognition. Iizuka et al. \cite{iizuka2017globally} proposed a generative model for face restoration. The proposed model employed an adversarial training approach using global and local context discriminators. The global discriminator assesses the entire image and the local discriminator looks at a small area in the completed region to ensure consistency with generated patches. An improvement to this approach was proposed by Yu et al. \cite{yu2018generative}. The improvement was to split the image completion network into a coarse network and a refinement network. The refinement network takes the initial coarse prediction and produces refined results. The authors used Wasserstein GAN (WGAN) \cite{arjovsky2017wasserstein} in their network to improve the results.

In contrast to several cases of image inpainting where the missing part is small and not complex in shape, a facial mask covers a big region of the face which makes this task more challenging. 
Din et al. \cite {din2020novel} proposed a generative network consisting of two discriminators to learn the general face shape and one generator. This approach was capable of removing the facial mask using the binary map and synthesizing the missing regions while keeping the initial face structure.
The proposed mask extraction encoder uses five blocks of convolution layers. The decoder component has the same architecture as the encoder except for the convolution layers that were replaced with deconvolution layers. 
This approach was evaluated on CelebA Dataset and structural similarity (SSIM) of 0.864 was reported.

Yu et al. \cite{yu2019free} used a gated convolutional network that provides a learnable dynamic feature selection mechanism for each channel at each spatial location across all layers. In addition, patch-based GAN (SN-PatchGAN) was used in their approach which contributed to reducing the loss value. LU et al. \cite{lu2022diverse} proposed an EXE-GAN network to reconstruct the face with an exemplar image without changing the color of known regions. The inpainting framework consists of a mapping network, a style encoder, a multi-style generator, and a discriminator. The proposed approach was evaluated on a free-form masked CelebA Dataset and a Frechet inception distance (FID) of 4.750 was reported. Ge et al. \cite{ge2020occluded} proposed an ID-GAN model for recognizing masked faces. The proposed model consists of an integrated CNN-based face recognizer with a generative adversarial network.  
 


\section{Proposed Approach}
\label{sec:proposed_approach}

Several models have been proposed in this work for masked face recognition. The input to the proposed system is a face image with a mask and the output is the identity of the person. We categorize the proposed models into fine-tuned, Transformer, and ensemble learning based models. 

\subsection{Fine-tuned Models}
Convolutional neural networks (CNN) is one of the most commonly used classification models in pattern recognition. This model is efficient in extracting the spatial features in still images \cite{luqman2021joint}. To utilize the capabilities of CNN for our task and to avoid training the model from scratch, we proposed three CNN models fine-tuned on different pre-trained model.  


The first CNN model was fine-tuned using the VGG16 model \cite{simonyan2014very}. VGG16 is one of the earliest architectures that has been used for image classification. This model contributed to the fast improvement of several deep learning models for different classification problems. We used VGG16 to extract the spatial features from the input images, and the output features vector of size 512 was fed into three layers. The first layer is a dropout layer with a 0.5 probability used to reduce the possibility of overfitting. The output of this layer was fed into a batch normalization layer followed by a classification layer.

We also proposed a CNN model by fine-tuning the EfficientNet pre-trained model. EfficientNet is a CNN model that depends on component scaling to boost the model performance with a few numbers of parameters \cite{tan2019efficientnet}. The scaling method uniformly scales all dimensions of the input using a compound coefficient. An EfficientNet-B0 model is used in this work to extract features from the input images and we fed these features into a dropout layer followed by a batch normalization layer. The output of this layer is fed into the classification layer.

The third CNN model that has been proposed in this work utilizes FaceNet pre-trained model. FaceNet was proposed by Florian Schroff et al. \cite{schroff2015facenet} at Google in 2015 for face recognition. In contrast to VGG16 and EfficientNet pre-trained models that were trained on the ImageNet dataset, FaceNet was trained on face dataset. 
FaceNet is a deep CNN model trained via a triplet loss function that encourages embedding vectors for the same person to become more similar by having smaller Euclidean distances compared with vectors for different persons who are expected to have larger distances. We fine-tuned this model for our task since the model was trained on persons' faces without masks, whereas our dataset contains both masked and unmasked faces. Therefore, we fine-tuned the last layers of this model and added a dropout and batch normalization layers to this model. The output features of this model were fed into the classification layer.  We used a Softmax function in the classification layer for all CNN models. The Softmax function assigns a probability for each predicted subject. The number of neurons in the classification layer matches the number of subjects in the dataset.


\subsection{Transformer}

The transformer is a deep learning model that learns spatial and temporal information by tracking relationships between each part of the input of sequential data \cite{vaswani2017attention}. This model depends on a self-attention mechanism to detect the significance of each part of the series. Transformer was proposed primarily for natural language processing problems such as machine translation and text summarizing. However, Transformer has been used recently for other fields and it became a dominant model for vision tasks.

The original architecture of the Transformer model consists of two main components, encoder and decoder. The encoder transforms the input data series into a hidden state to be used by the decoder component that is responsible for output generation. In this work, we proposed a Transformer model for masked face recognition that uses only the encoder component since this problem is a classification problem that does not require the decoder component. This architecture is followed by several recent models that employ Transformer for classification tasks such as BERT model \cite{devlin2018bert} which was proposed for several natural language processing tasks.  
 
\begin{figure*}[!t]
    \centering
	\includegraphics[width=0.9\textwidth]{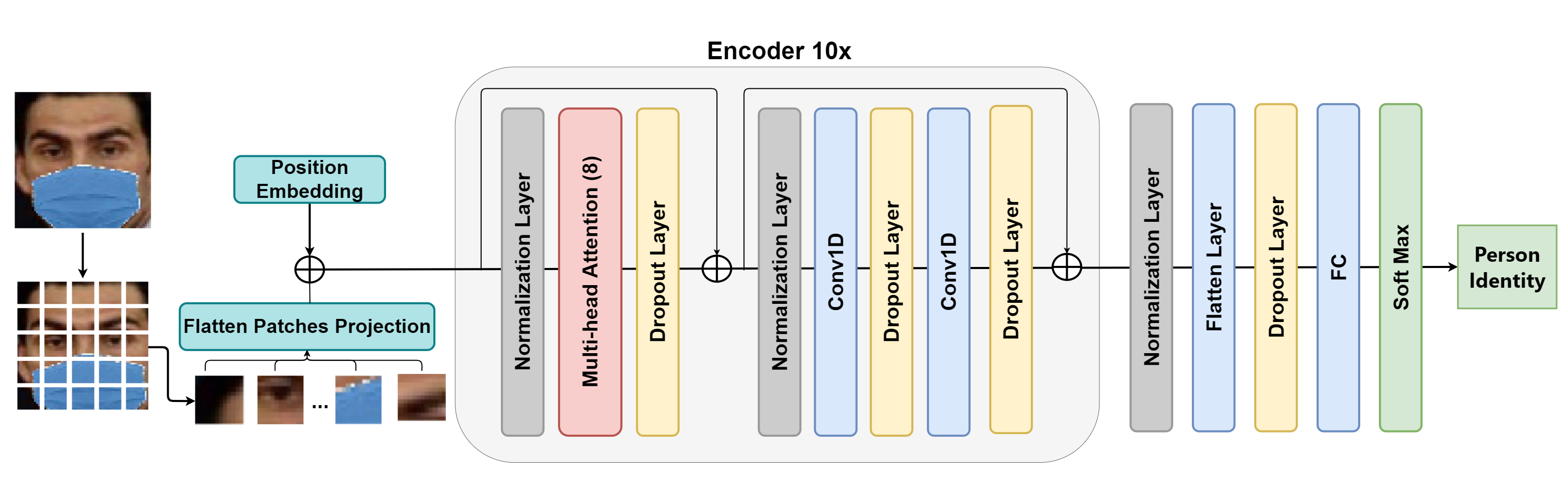}
    \caption{The architecture of the Transformer model.}
    \label{fig:transformer}
    \end{figure*}

The proposed Transformer which was inspired by the design of Dosovitskiy et al. \cite{dosovitskiy2020image} accepts an input image and splits it into a number of patches as shown in Figure \ref{fig:transformer}. Each patch is fed into the encoder representing one time step. However, this image segmentation does not preserve the spatial information of the input image. To address this issue, a position embedding is used with each patch to encode its position in the original input image. The resulting patch with its embedded position is fed into the Transformer's encoder. The encoder consists of a layer normalization followed by multi-head attention (MHA) unit. A residual connection is also used with MHA unit and the output of the connection is fed into another normalization layer followed by a multi-perception layer (MPL). MPL consists of two blocks of fully connected (FC) and dropout layers with different configurations. The output of the MPL is added with the MHA output to form the output of the transformer. The different hyper-parameters used with the Transformer are shown in table \ref{tab:Transformer_param}.

\begin{table}[H]
\centering
\footnotesize
\caption{Transformer model hyper-parameters.}
\label{tab:Transformer_param}
\setlength{\tabcolsep}{12pt}
\begin{tabular}{@{}ll@{}}
\toprule
Hyper-parameter       &   Value        \\ \midrule
\# Transformer Blocks &   10            \\
MHA Key Dimension     &   64           \\
Number of heads       &   8            \\
Dropout in Encoder      &   0.3          \\
Units in the FC Layer   &   [2048, 1024] \\ 
Dropout probability   &   0.6    \\ 
\bottomrule
\end{tabular}
\end{table}

\subsection{Ensemble Learning}

We employed ensemble learning in this work for masked face recognition. Ensemble learning is an efficient machine learning approach that seeks better predictive performance by combining the predictions from multiple models \cite{rokach2010ensemble}. 
Two approaches are used in this work for ensemble learning. The first method involves using different validation sets for each model. This method helps in overcoming the overfitting problem of the involved models. The second method is to ensemble different models with different configurations.  This helps in employing the capabilities of these models for masked face identification. In addition, combining multiple models’ predictions can reduce the variance of the resulting ensembled model. 

\begin{figure}[h] 
    \centering
	\includegraphics[width=\linewidth]{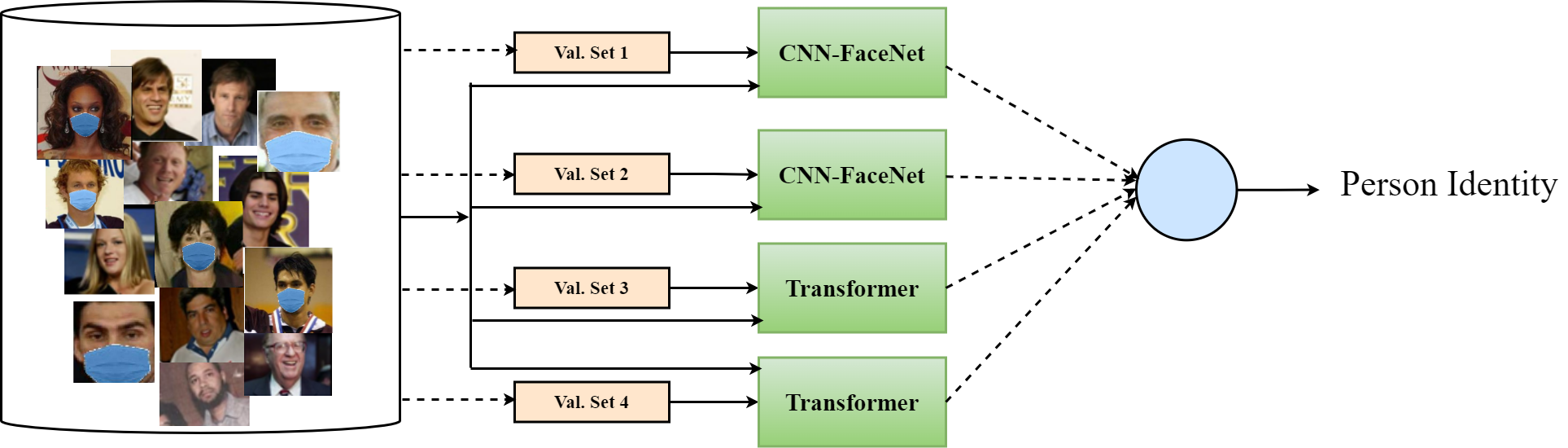}
    \caption{The framework of the ensemble learning model.}  
    \label{fig:ensemble}
    \end{figure}

The proposed model combines the predictions of two convolutional networks and two Transformer models as shown in Figure \ref{fig:ensemble}. The two CNN models are fined-tuned on FaceNet pre-trained model. Each of the ensembled models is trained and validate on a different set of the dataset. Majority voting technique is used to find the final prediction of the model by considering the prediction of all participated models.

\section{Experimental Work}

\subsection{Dataset}

\begin{figure}[h]
    \centering
    \subfloat[\centering LFW dataset ]{{\includegraphics[width=0.8\linewidth]{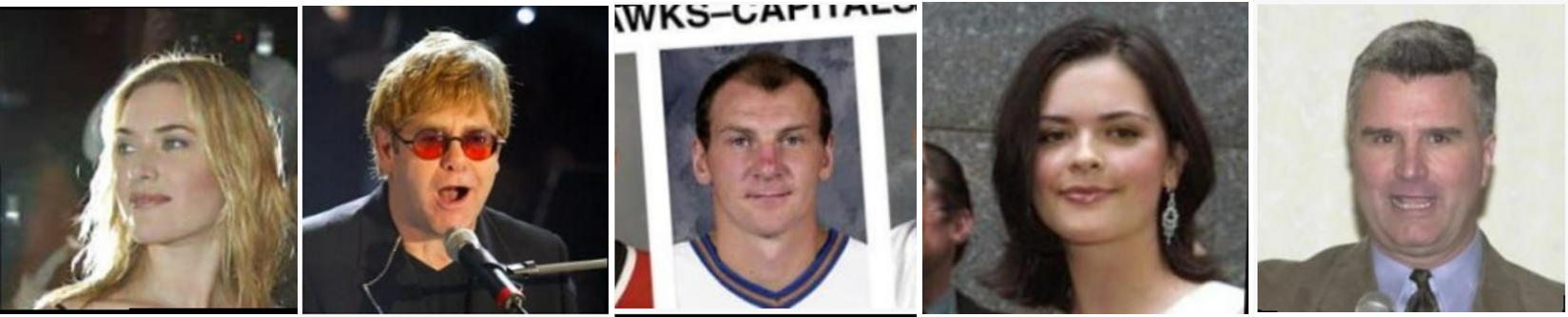} }}%
     \qquad
    \subfloat[\centering Masked LFW  ]{{\includegraphics[width=0.8\linewidth]{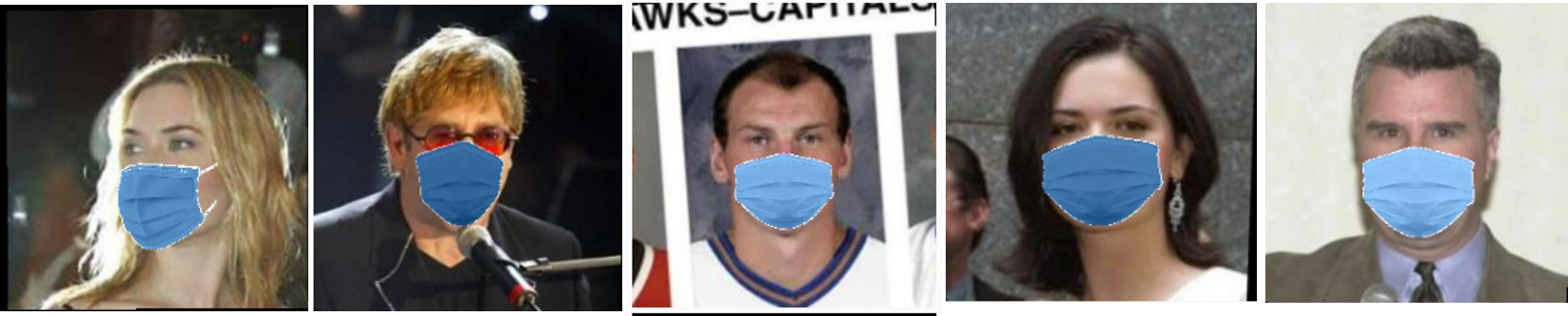}} }%
    \caption{Samples of the LFW dataset and the generated masked faces.}  
    \label{fig:dataset}%
\end{figure}

The labeled Faces in the Wild (LFW) dataset is used in this work to train and evaluate the proposed models. The dataset consists of 13,233 images of faces collected from the web. All the images were captured in an unconstrained environment. The number of subjects in the dataset is 5,749 with 1,680 subjects having two or more pictures. Samples of this dataset are shown in Figure \ref{fig:dataset} (a).

LFW dataset does not contain faces with masks. To address this issue, we propose a masked face dataset created from the LFW dataset. MaskTheFace model \cite{anwar2020masked} is used in this work to create a version of LFW with masked faces. This model utilizes a face landmarks detector to detect six key points on the face necessary for applying the mask.  The surgical mask is then transformed based on these six key points to fit perfectly on the face. Samples of the dataset with masks are shown in Figure \ref{fig:dataset} (b).

\subsection{Results and Discussion}
Five models have been proposed in this work for masked face recognition. Three of these models depend on fine-tuning three different pre-trained models, namely VGG16, EfficientNet, and FaceNet. A transformer model is also proposed in this work for masked face identification. To utilize the features of the fine-tuned models and the Transformer model, we also propose an ensemble learning model that combines four models with different configurations and setups. More details about these models are presented in Section \ref{sec:proposed_approach}.

The proposed models have been trained and evaluated on the LFW dataset and another version of this dataset containing the same subjects of the LFW dataset with face masks. The dataset consists of 26,466 images representing 5,749 subjects. These images are randomly split into 95\% for training and 5\% for testing using a random seed of 777.   
However, this dataset is imbalanced and it contains 4,047 out of 5,749 subjects with only one image per subject. To address this issue, we performed data augmentation. Three transformations have been applied to the training samples. These augmentation transformations involve horizontal flipping, rescaling, and zooming. These operations have been selected based on the possible variations that may happen when acquiring the person's image. For example, the position of the camera may be close or far from the person, or the subject may not be always facing the camera.    

Table \ref{tab:results} shows the obtained results using the proposed models. We report the top-1 and top-5 accuracies of each model. As shown in the table, the fine-tuned CNN model on FaceNet obtained the highest recognition accuracy compared with other CNN models. This can be attributed to the datasets that have been used to train these models. FaceNet model was trained on face images whereas VGG16 and EfficientNet were trained on the ImageNet dataset that contains different classes. This justifies the low accuracies obtained using VGG16 and EfficientNet models.  

Although the Transformer model showed a good performance for several computer vision problems in the literature, the obtained result in this work is less than the fine-tuned models. This can be attributed to the dataset size that is used to train the Transformer model. We used only around 25K images for model training whereas the number of subjects in this dataset is 5749 subjects. In addition, the model accuracy was penalized by the nature of the LFW dataset which contains 4,047 out of 5,749 subjects with only one image.
To address this issue and utilize the pre-trained models for masked face recognition, ensemble learning is used. As shown in Table \ref{tab:results}, the ensemble learning model outperformed other models with a top-1 accuracy of 92.01\% and a top-5 accuracy of 96.57\%. These results show that combining the CNN-FaceNet model with the Transformer models tackles the dataset size issue of the Transformer model and utilizes the pre-trained models for boosting the recognition rate.

\begin{table}[H]
\centering
\footnotesize
\caption{The accuracies of the proposed models.}
\label{tab:results}
\setlength{\tabcolsep}{12pt}
\begin{tabular}{@{}lll@{}}
\toprule
Model       &   Top-1 Accuracy     &   Top-5 Accuracy       \\ \midrule
CNN-VGG16 &    73.38\%    &  82.05\%      \\
CNN-EfficientNet     &   79.61\%   &  84.41\%        \\
CNN-FaceNet       &     80.30\%    &   85.24\%    \\
Transformer      &       69.04\%  &   78.70\%  \\
Ensemble Learning  &  \textbf{92.01\%} & \textbf{96.57\%} \\ 
\bottomrule
\end{tabular}
\end{table} 
 
\begin{figure}[H] 
    \centering
	\includegraphics[width=0.75\linewidth]{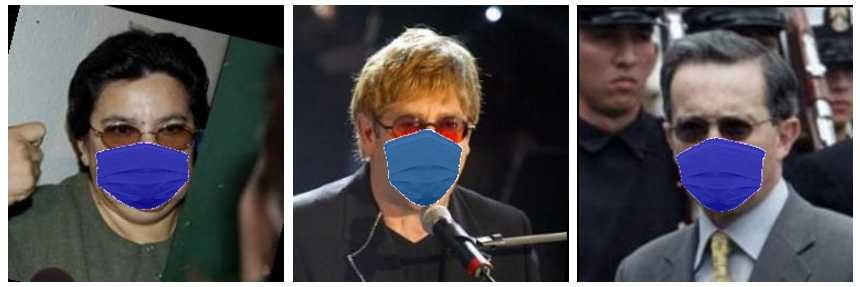}
    \caption{Misclassified samples.}  
    \label{fig:misclassification}
    \end{figure}

The misclassification of the masked faces can be attributed mainly to the dataset. The dataset used in this work is imbalanced with few samples per subject. Based on the error analysis that we conducted, 52\% of the misclassified samples were for subjects that have only one sample without a mask in the train data. This number of samples per class is not enough to train the proposed models to recognize the same subject with a mask. Another source of error was with persons whose facial expressions are not shown when they are wearing masks. This happens usually with people who are wearing sunglasses as shown in Figure \ref{fig:misclassification}. This type of samples were difficult to recognize given the limited number of training samples and the lack of enough visual clues.

\section{Conclusions and Future Work}

Wearing face masks is one of the preventive measures followed to avoid the spread of coronavirus. This procedure became mandatory in several countries. 
Although the face recognition systems have now advanced to the point of being commercially available, these systems do not generalize well with faces occluded with masks. To address this problem, we propose a masked face recognition system. Five models have been proposed in this work. Three of these models are CNN models fine-tuned on three pre-trained models, namely VGG16, EfficientNet, and FaceNet. In addition, we proposed a Transformer based model for masked face recognition. We also proposed an ensemble learning model by combining convolutional and Transformer models.  
The proposed models have been trained and evaluated on a synthetically masked LFW dataset that has been proposed in this work. The obtained results show that the CNN model fine-tuned using FaceNet pre-trained model outperformed other CNN models. In addition, the model created using ensemble learning outperformed other models with an accuracy of 92.01\%. These results show that our proposed model is efficient for masked face recognition. In future work, we will work on handling the problem of misclassifying masked faces of people wearing sunglasses. We will also address the dataset size limitation and evaluate the proposed models on masked face verification.

\section*{Acknowledgment}

The authors would like to acknowledge the support provided by King Fahd University of Petroleum and Minerals (KFUPM) during this work.

\bibliographystyle{IEEEtran}

\end{document}